\documentclass{article}

\usepackage[preprint]{neurips_2026}

\usepackage[utf8]{inputenc}
\usepackage[T1]{fontenc}
\usepackage{hyperref}
\usepackage{url}
\usepackage{booktabs}
\usepackage{amsfonts}
\usepackage{amsmath}
\usepackage{amssymb}
\usepackage{nicefrac}
\usepackage{microtype}
\usepackage{xcolor}
\usepackage{graphicx}
\usepackage{algorithm}
\usepackage{algorithmic}
\usepackage{multirow}
\usepackage{subcaption}
\usepackage{wrapfig}
\usepackage{amsthm}
\newtheorem{theorem}{Theorem}
\newtheorem{lemma}[theorem]{Lemma}

\newcommand{\ours}{\textsc{3DTurboQuant}}

\newcommand{\R}{\mathbb{R}}
\newcommand{\E}{\mathbb{E}}
\newcommand{\Var}{\mathrm{Var}}
\newcommand{\bx}{\boldsymbol{x}}
\newcommand{\by}{\boldsymbol{y}}

\newcommand{\bK}{\boldsymbol{K}}
\newcommand{\bV}{\boldsymbol{V}}
\newcommand{\bQ}{\boldsymbol{Q}}

\newcommand{\bT}{\boldsymbol{T}}
\newcommand{\bPi}{\boldsymbol{\Pi}}
\newcommand{\bmu}{\boldsymbol{\mu}}
\newcommand{\bq}{\boldsymbol{q}}
\newcommand{\bs}{\boldsymbol{s}}
\newcommand{\ip}[2]{\langle #1, #2 \rangle}

\title{\ours{}: Training-Free Near-Optimal Quantization for 3D Reconstruction Models}

\author{
  Jae Joong Lee\\  Department of Computer Science\\
  Purdue University\\
  \texttt{lee2161@purdue.edu} \\
}

\begin{document}

\maketitle

\begin{abstract}
Every existing method for compressing 3D Gaussian Splatting, NeRF, or transformer-based 3D reconstructors requires learning a data-dependent codebook through per-scene fine-tuning. We show this is unnecessary. The parameter vectors that dominate storage in these models, 45-dimensional spherical harmonics in 3DGS and 1024-dimensional key-value vectors in DUSt3R, fall in a dimension range where a single random rotation transforms any input into coordinates with a known Beta distribution. This makes precomputed, data-independent Lloyd-Max quantization near-optimal, within a factor of 2.7 of the information-theoretic lower bound. We develop \ours{}, deriving (1) a dimension-dependent criterion that predicts which parameters can be quantized and at what bit-width before running any experiment, (2) norm-separation bounds connecting quantization MSE to rendering PSNR per scene, (3) an entry-grouping strategy extending rotation-based quantization to 2-dimensional hash grid features, and (4) a composable pruning-quantization pipeline with a closed-form compression ratio. On NeRF Synthetic, \ours{} compresses 3DGS by \textbf{3.5$\times$ with 0.02\,dB PSNR loss} and DUSt3R KV caches by \textbf{7.9$\times$ with 39.7\,dB pointmap fidelity}. No training, no codebook learning, no calibration data. Compression takes seconds.
\end{abstract}

\section{Introduction}
\label{sec:intro}

Compressing 3D reconstruction models today requires training. For 3D Gaussian Splatting (3DGS)~\cite{kerbl3Dgaussians}, methods like CompGS~\cite{navaneet2024compgs}, HAC++~\cite{chen2025hacpp}, and OMG~\cite{lee2025omg} learn per-scene codebooks through hours of fine-tuning to reach 20--185$\times$ compression. For NeRF~\cite{mildenhall2020nerf, muller2022instant} hash grids, SHACIRA~\cite{girish2023shacira} and CNC~\cite{chen2024cnc} train entropy models per scene. For transformer reconstructors like DUSt3R~\cite{wang2024dust3r}, KV cache quantization methods~\cite{liu2024kivi, hooper2024kvquant} require calibration data. Every method in every 3D reconstruction family shares the same structural requirement: a data-dependent codebook or calibration step that must be repeated for each new scene or model.

This requirement has practical consequences. A streaming 3D application cannot pause to fine-tune a codebook. A dynamic scene with densifying Gaussians invalidates codebooks learned on earlier states. An on-device deployment cannot afford the GPU-hours needed for per-scene compression. The question is whether data-dependent codebook learning is fundamentally necessary, or whether the structure of 3D reconstruction parameters admits a data-independent alternative.

We find that it is not necessary. The parameter vectors that dominate storage in 3D reconstruction models occupy a specific dimension range, $d \in [16, 1024]$, where rotation-based vector quantization~\cite{zandieh2025turboquant} achieves near-optimal distortion \emph{without any data-dependent learning}. The mechanism is the following: multiplying a $d$-dimensional vector by a random orthogonal matrix produces coordinates that follow a Beta distribution with variance $1/d$. When $d$ is large enough (we find $d \geq 16$ suffices in practice), these coordinates are nearly independent, and a \emph{precomputed} Lloyd-Max scalar quantizer~\cite{lloyd1982least} for the Beta distribution is near-optimal. 3DGS spherical harmonic coefficients have $d=45$. DUSt3R KV cache vectors have $d=1024$. Both fall squarely in this range.

Building on this observation, we make four contributions:

\begin{enumerate}
    \item \textbf{Dimension-dependent quantization criterion.} We derive which 3D reconstruction parameters can be quantized by rotation-based VQ based on their dimension $d$, and at what bit-width $b$. We show that coordinates at $d=45$ are independent enough for near-optimal scalar quantization at $b \geq 3$, while $d=3$ (positions) and $d=4$ (quaternions) are not. This criterion predicts per-bit rendering PSNR loss \emph{before any experiment}: at $b=3$ and $d=45$, the bound gives $D_\text{mse} \leq 0.03$, and we measure $0.033$ on Lego, a 10\% gap.

    \item \textbf{Norm-separation bounds.} 3D reconstruction parameters are not unit-norm, unlike the setting analyzed in~\cite{zandieh2025turboquant}. We derive that separating the norm $\gamma_i = \|\boldsymbol{f}_i\|_2$ and quantizing the direction $\hat{\boldsymbol{f}}_i = \boldsymbol{f}_i / \gamma_i$ yields per-element MSE of $\gamma_i^2 \cdot \frac{\sqrt{3\pi}}{2} \cdot 4^{-b}$. This gives a closed-form prediction of rendering quality as a function of bit-width and the SH norm distribution of each scene.

    \item \textbf{Entry-grouping for low-dimensional features.} Instant-NGP~\cite{muller2022instant} hash entries have $d_f = 2$, below the threshold where coordinate independence holds. We introduce a grouping strategy that concatenates $g$ entries into $d_\text{eff} = g \cdot d_f$ dimensions before rotation and quantization, extending the approach to NeRF feature grids.

    \item \textbf{Composable compression with derived rates.} We show that rotation-based quantization composes multiplicatively with opacity pruning (retaining fraction $\rho$) and SH degree reduction (factor $r$), with a closed-form total compression ratio of $\frac{1}{\rho} \cdot \frac{32}{b \cdot r + 56/d_\text{sh}}$. This yields 5--8$\times$ total compression on 3DGS without any retraining.
\end{enumerate}

\section{Related Work}
\label{sec:related}

\paragraph{3D Gaussian Splatting compression.}
The growing memory cost of 3DGS has motivated a rich line of compression work, recently surveyed in~\cite{bagdasarian2025survey}. Methods can be broadly categorized into three strategies that are typically combined.
\emph{Codebook-based quantization:} CompGS~\cite{navaneet2024compgs} trains a VQ-VAE to learn compact codebooks for Gaussian attributes with entropy coding (31$\times$). C3DGS~\cite{niedermayr2024c3dgs} applies sensitivity-aware vector clustering with quantization-aware training. Compact-3DGS~\cite{lee2024compact3dgs} replaces SH with a grid-based neural field and applies codebook VQ (25$\times$+).
\emph{Context and entropy modeling:} HAC~\cite{chen2024hac} introduces hash-grid-assisted spatial context models with arithmetic coding. Its extension HAC++~\cite{chen2025hacpp} achieves over 100$\times$ compression by explicitly minimizing entropy during optimization. ContextGS~\cite{wang2024contextgs} develops anchor-level autoregressive context models (20$\times$). CodecGS~\cite{lee2025codecgs} maps Gaussians to tri-plane feature planes and leverages standard video codecs (H.265/VVC) for 146$\times$ compression.
\emph{Pruning and distillation:} LightGaussian~\cite{fan2024lightgaussian} combines global significance pruning with SH distillation and VecTree quantization (15$\times$). LP-3DGS~\cite{zhang2024lp3dgs} learns differentiable pruning masks. EAGLES~\cite{girish2024eagles} uses quantized embeddings with progressive training. SOGS~\cite{morgenstern2024sogs} arranges Gaussians into a 2D grid for off-the-shelf image codec compression (17--42$\times$).

Wang et al.~\cite{wang2025noisevq} propose noise-substituted VQ that jointly trains codebooks and features ($\sim$45$\times$). SALVQ~\cite{xu2025salvq} replaces uniform scalar quantization with scene-adaptive lattice VQ. A common thread across all these methods is their reliance on \emph{data-dependent, per-scene training}: codebooks, context models, and entropy parameters must be learned anew for each scene, typically taking hours. The sole exception is FlexGaussian~\cite{tian2025flexgaussian}, which is training-free but uses heuristic mixed-precision assignment without theoretical guarantees. \ours{} provides the quantization component with \emph{provable near-optimality} using a fixed, precomputed codebook, bridging the gap between training-free convenience and theoretically-grounded compression.

\paragraph{Neural Radiance Field compression.}
NeRF compression targets the learned feature representations that dominate storage. SHACIRA~\cite{girish2023shacira} develops importance-weighted hash-grid codebooks with quantization-aware retraining for Instant-NGP. CNC~\cite{chen2024cnc} exploits level-wise and dimension-wise context dependencies in hash grids, achieving 100$\times$ compression on NeRF Synthetic. VQRF~\cite{li2023vqrf} applies vector quantization to TensoRF~\cite{chen2022tensorf} factored features. VQAD~\cite{takikawa2022vqad} proposes a vector-quantized auto-decoder for variable-bitrate neural fields. More recently, HERO~\cite{zhang2025hero} introduces RL-based hardware-aware quantization for NeRF accelerators, Quant-NeRF~\cite{hasssan2025quantnerf} develops end-to-end quantization for low-precision 3D Gaussian NeRF, and Zhang et al.~\cite{zhang2025posenc} propose hardware-friendly positional encoding quantization. All are data-dependent: codebooks or quantization parameters must be learned per scene. \ours{} applies a \emph{fixed, precomputed} codebook derived from the Beta distribution, avoiding any per-scene learning.

\paragraph{KV cache quantization for transformers.}
Memory-efficient inference in transformers has driven work on KV cache compression, both for LLMs and emerging 3D vision transformers. KIVI~\cite{liu2024kivi} proposes per-channel asymmetric 2-bit quantization. KVQuant~\cite{hooper2024kvquant} uses sensitivity-weighted quantization with per-channel scales. QJL~\cite{zandieh2024qjl} introduces a 1-bit scheme based on the Johnson-Lindenstrauss transform providing unbiased inner product estimation. PolarQuant~\cite{han2025polarquant} decomposes vectors using polar coordinates. For 3D vision transformers specifically, QuantVGGT~\cite{feng2025quantvggt} applies W4A4 post-training quantization to the 1.2B-parameter VGGT model with Hadamard rotation smoothing. XStreamVGGT~\cite{xstreamvggt2026} combines token-importance pruning with dimension-adaptive KV quantization for 4.4$\times$ memory reduction. TurboQuant~\cite{zandieh2025turboquant} extends these ideas with provably optimal MSE bounds by exploiting the Beta distribution of randomly-rotated coordinates. Our work applies this approach to DUSt3R, demonstrating that provably near-optimal quantization achieves 7.9$\times$ KV compression with high-fidelity 3D reconstruction.

\paragraph{Vector quantization theory.}
The information-theoretic foundation for vector quantization was laid by Shannon's distortion-rate theory~\cite{shannon1948, shannon1959}, establishing that the minimum achievable distortion for a source with differential entropy $h(\bx)$ at bit budget $B$ is $D(B) \geq \frac{d}{2\pi e} \cdot 2^{(2/d)(h(\bx) - B)}$. Zador~\cite{zador1964} derived asymptotic expressions for fixed-rate quantizers, and Gersho~\cite{gersho1979} popularized lattice quantization. The Lloyd-Max algorithm~\cite{lloyd1982least, max1960quantizing} provides the optimal scalar quantizer for known distributions. TurboQuant~\cite{zandieh2025turboquant} achieves the Shannon bound within a constant factor by exploiting the fact that random rotation transforms worst-case inputs into vectors with a known, quantization-friendly distribution.

\section{Preliminaries}
\label{sec:prelim}

We first establish the formal problem definition, then briefly review the three 3D reconstruction settings and the TurboQuant algorithm that underlies our approach.

\subsection{Problem Definition}
\label{sec:problem}

Let $\Theta = \{\boldsymbol{\theta}_1, \ldots, \boldsymbol{\theta}_N\} \subset \R^d$ denote the set of $N$ parameter vectors of dimension $d$ in a trained 3D reconstruction model. Our goal is to design a quantization scheme that compresses each $\boldsymbol{\theta}_i$ from $32d$ bits (float32) to $bd$ bits ($b$ bits per coordinate, $b \ll 32$), while minimizing the distortion in the model's output.

Formally, we seek a quantization map $Q : \R^d \to \{0,1\}^{bd}$ and dequantization map $Q^{-1} : \{0,1\}^{bd} \to \R^d$ that minimize the worst-case expected MSE distortion:
\begin{equation}
    D_\text{mse} := \max_{\bx \in \mathbb{S}^{d-1}} \E_Q\left[\left\|\bx - Q^{-1}(Q(\bx))\right\|_2^2\right],
    \label{eq:mse_def}
\end{equation}
where the expectation is over the randomness in $Q$ (which may be a randomized quantizer) and the maximization is over all unit-norm input vectors.

For applications involving inner product computation (e.g., attention in transformers), we also consider the inner product distortion:
\begin{equation}
    D_\text{prod} := \max_{\substack{\bx \in \mathbb{S}^{d-1} \\ \by \in \R^d}} \E_Q\left[\left|\ip{\by}{\bx} - \ip{\by}{Q^{-1}(Q(\bx))}\right|^2\right],
    \label{eq:prod_def}
\end{equation}
with the additional desideratum of \emph{unbiasedness}: $\E_Q\left[\ip{\by}{Q^{-1}(Q(\bx))}\right] = \ip{\by}{\bx}$.

\textbf{Design requirements.} For 3D reconstruction deployment, the quantizer must satisfy three properties beyond low distortion: (i)~\emph{data-oblivious}: no access to the training data or calibration set. (ii)~\emph{online}: each vector is quantized independently, enabling streaming and dynamic scenes. (iii)~\emph{computationally efficient}: quantization should be faster than model training by orders of magnitude.

\subsection{3D Reconstruction Approaches}
\label{sec:approaches}

We briefly describe the parameter structures of each approach to motivate our quantization targets.

\paragraph{3D Gaussian Splatting (3DGS).} A 3DGS model~\cite{kerbl3Dgaussians} represents a scene as a set of $N$ anisotropic Gaussians $\{(\bmu_i, \boldsymbol{\Sigma}_i, \alpha_i, \boldsymbol{c}_i)\}_{i=1}^N$, where $\bmu_i \in \R^3$ is the center, $\boldsymbol{\Sigma}_i$ is the covariance (parameterized by scale $\bs_i \in \R^3$ and rotation quaternion $\bq_i \in \R^4$), $\alpha_i \in \R$ is the opacity (stored in logit space), and $\boldsymbol{c}_i$ is the view-dependent color encoded via spherical harmonics (SH). At SH degree $l$, the color is represented by DC coefficients $\boldsymbol{c}_i^\text{dc} \in \R^3$ and higher-order rest coefficients $\boldsymbol{f}_i \in \R^{d_\text{sh}}$ where $d_\text{sh} = 3((l+1)^2 - 1)$. For $l=3$: $d_\text{sh} = 45$, constituting 180 of the 236 bytes per Gaussian (\textbf{76\%}).

\paragraph{Neural Radiance Fields (NeRF).} Instant-NGP~\cite{muller2022instant} encodes scene geometry and appearance in multi-resolution hash tables $\{\bT^{(r)}\}_{r=1}^R$ across $R$ resolution levels. Each table $\bT^{(r)} \in \R^{N_r \times d_f}$ stores $N_r$ feature vectors of dimension $d_f$ (typically $d_f = 2$). A query point $\bx \in \R^3$ is projected to each level, the enclosing voxel vertices are looked up, features are trilinearly interpolated, concatenated across levels, and passed through a small MLP to produce density and color.

\paragraph{Transformer reconstruction (DUSt3R).} DUSt3R~\cite{wang2024dust3r} uses a ViT-Large encoder~\cite{dosovitskiy2020vit} with $L$ transformer layers, each with multi-head self-attention over $H$ heads of dimension $d_h$. For $V$ input views tokenized into $P$ patches each, the self-attention at layer $\ell$ computes:
\begin{equation}
    \text{Attn}^{(\ell)} = \text{softmax}\!\left(\frac{\bQ^{(\ell)} {\bK^{(\ell)}}^\top}{\sqrt{d_h}}\right) \bV^{(\ell)},
    \label{eq:attn}
\end{equation}
where $\bQ^{(\ell)}, \bK^{(\ell)}, \bV^{(\ell)} \in \R^{VP \times d_\text{kv}}$ with $d_\text{kv} = H \cdot d_h$. The KV cache for all layers consumes $2L \cdot VP \cdot d_\text{kv} \cdot 4$ bytes. For DUSt3R ViT-Large ($L=24$, $H=16$, $d_h=64$, $d_\text{kv}=1024$), this grows to hundreds of MB for multi-view inputs.

\subsection{TurboQuant: Near-Optimal Data-Oblivious Quantization}
\label{sec:turboquant}

TurboQuant~\cite{zandieh2025turboquant} solves the problem in Eq.~\eqref{eq:mse_def} by reducing vector quantization in $\R^d$ to $d$ independent scalar quantization problems. The key enabling result is the following:

\begin{lemma}[Coordinate distribution on the hypersphere~\cite{zandieh2025turboquant}]
\label{lem:beta}
If $\bx \in \mathbb{S}^{d-1}$ is uniformly distributed on the unit hypersphere, then each coordinate $\bx_j$ follows the Beta distribution:
\begin{equation}
    \bx_j \sim f_X(x) := \frac{\Gamma(d/2)}{\sqrt{\pi}\,\Gamma((d-1)/2)}\left(1 - x^2\right)^{(d-3)/2}, \quad x \in [-1, 1].
    \label{eq:beta}
\end{equation}
In high dimensions, $f_X(\cdot) \to \mathcal{N}(0, 1/d)$, and distinct coordinates become nearly independent.
\end{lemma}

Since multiplying any fixed $\bx \in \mathbb{S}^{d-1}$ by a random orthogonal matrix $\bPi$ (obtained via QR decomposition of an i.i.d.\ Gaussian matrix) produces $\by = \bPi \bx$ uniformly distributed on $\mathbb{S}^{d-1}$, each coordinate $\by_j$ follows $f_X$. This transforms the \emph{worst-case} vector quantization problem into one where the coordinate distribution is \emph{known}, enabling the use of optimal scalar quantization.

\paragraph{Optimal scalar quantization.} The Lloyd-Max algorithm~\cite{lloyd1982least, max1960quantizing} finds centroids $\{c_1, \ldots, c_{2^b}\}$ that minimize the scalar quantization MSE for a given distribution. For $f_X$ in Eq.~\eqref{eq:beta}, this amounts to solving:
\begin{equation}
    \mathcal{C}(f_X, b) := \min_{-1 \leq c_1 \leq \cdots \leq c_{2^b} \leq 1} \sum_{i=1}^{2^b} \int_{\frac{c_{i-1}+c_i}{2}}^{\frac{c_i+c_{i+1}}{2}} |x - c_i|^2 \cdot f_X(x)\, dx.
    \label{eq:kmeans}
\end{equation}
Crucially, this codebook depends only on $d$ and $b$. It can be precomputed once and reused for all scenes.

\paragraph{TurboQuant$_\text{mse}$ algorithm.} The complete procedure is:
\begin{enumerate}
    \item \textbf{Setup} (once per $(d, b)$): Generate random rotation $\bPi \in \R^{d \times d}$; compute codebook $\{c_k\}$ by solving Eq.~\eqref{eq:kmeans}.
    \item $\textsc{Quantize}(\bx)$: $\by \leftarrow \bPi \cdot \bx$;\; $\text{idx}_j \leftarrow \arg\min_k |\by_j - c_k|$ for $j \in [d]$;\; output idx.
    \item $\textsc{DeQuantize}(\text{idx})$: $\tilde{\by}_j \leftarrow c_{\text{idx}_j}$;\; $\tilde{\bx} \leftarrow \bPi^\top \cdot \tilde{\by}$;\; output $\tilde{\bx}$.
\end{enumerate}

\begin{theorem}[MSE bound~\cite{zandieh2025turboquant}]
\label{thm:mse}
For any $b \geq 1$ and $\bx \in \mathbb{S}^{d-1}$, TurboQuant$_\text{mse}$ achieves:
\begin{equation}
    D_\text{mse}(Q_\text{mse}) \leq \frac{\sqrt{3\pi}}{2} \cdot \frac{1}{4^b} \approx \frac{2.72}{4^b}.
    \label{eq:mse_bound}
\end{equation}
For $b = 1,2,3,4$: $D_\text{mse} \approx 0.36, 0.117, 0.03, 0.009$ respectively.
\end{theorem}

\begin{theorem}[Information-theoretic lower bound~\cite{zandieh2025turboquant}]
\label{thm:lower}
For \emph{any} randomized quantizer $Q : \mathbb{S}^{d-1} \to \{0,1\}^{bd}$ with any reconstruction map, there exist hard instances such that:
\begin{equation}
    D_\text{mse}(Q) \geq \frac{1}{4^b}, \qquad D_\text{prod}(Q) \geq \frac{\|\by\|_2^2}{d} \cdot \frac{1}{4^b}.
    \label{eq:lower}
\end{equation}
\end{theorem}

The ratio between the upper bound~\eqref{eq:mse_bound} and lower bound~\eqref{eq:lower} is $\frac{\sqrt{3\pi}}{2} \approx 2.7$, establishing \emph{near-optimality} of TurboQuant within a small constant factor of the best achievable distortion by \emph{any} algorithm.

\section{Method: \ours{}}
\label{sec:method}

\begin{figure}[t]
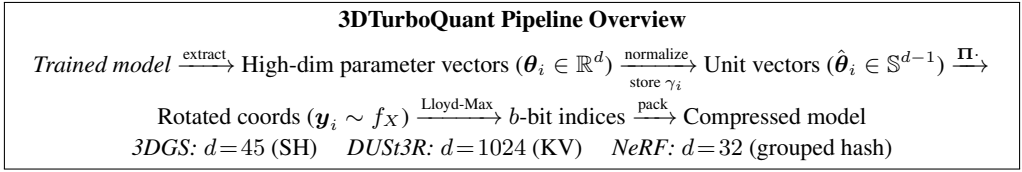

\centering
\fbox{\parbox{0.95\textwidth}{\centering\small
\textbf{3DTurboQuant Pipeline Overview}\\[4pt]
\textit{Trained model} $\xrightarrow{\text{extract}}$ High-dim parameter vectors ($\boldsymbol{\theta}_i \in \R^d$)
$\xrightarrow[\text{store } \gamma_i]{\text{normalize}}$ Unit vectors ($\hat{\boldsymbol{\theta}}_i \in \mathbb{S}^{d-1}$)
$\xrightarrow{\bPi \cdot}$ Rotated coords ($\by_i \sim f_X$)
$\xrightarrow{\text{Lloyd-Max}}$ $b$-bit indices
$\xrightarrow{\text{pack}}$ Compressed model\\[2pt]
\textit{3DGS:} $d\!=\!45$ (SH) \quad \textit{DUSt3R:} $d\!=\!1024$ (KV) \quad \textit{NeRF:} $d\!=\!32$ (grouped hash)
}}
\caption{Overview of \ours{}. Parameter vectors from any 3D reconstruction model are normalized, randomly rotated, and scalar-quantized using a precomputed codebook. The same algorithm applies across all three approaches. Only the dimension $d$ differs.}
\label{fig:overview}
\end{figure}

The core question is: given a trained 3D model with parameter vectors of dimension $d$, which vectors should be quantized, at how many bits, and how does the resulting distortion affect the model's output? We answer this for each of the three reconstruction approaches. The overall pipeline is shown in Figure~\ref{fig:overview}.

\subsection{3DGS Spherical Harmonic Compression}
\label{sec:method_3dgs}

For each Gaussian $i \in [N]$, we extract the SH rest coefficients as a flat vector $\boldsymbol{f}_i \in \R^{d_\text{sh}}$ ($d_\text{sh} = 45$ for $l=3$). We apply TurboQuant with norm separation:
\begin{equation}
    \gamma_i = \|\boldsymbol{f}_i\|_2, \qquad \hat{\boldsymbol{f}}_i = \boldsymbol{f}_i / \gamma_i, \qquad \text{idx}_i = \textsc{Quantize}(\hat{\boldsymbol{f}}_i).
\end{equation}
By Theorem~\ref{thm:mse}, the per-Gaussian SH reconstruction MSE is bounded:
\begin{equation}
    \E\left[\|\boldsymbol{f}_i - \tilde{\boldsymbol{f}}_i\|_2^2\right] \leq \gamma_i^2 \cdot \frac{\sqrt{3\pi}}{2} \cdot \frac{1}{4^b}.
    \label{eq:sh_mse}
\end{equation}

\paragraph{What we quantize vs.\ what we keep.} Positions $\bmu_i \in \R^3$, quaternions $\bq_i \in \R^4$, scales $\bs_i \in \R^3$, opacity $\alpha_i \in \R$, and DC color $\boldsymbol{c}_i^\text{dc} \in \R^3$ remain in float32. These low-dimensional parameters ($d \leq 4$) contribute only 56 bytes per Gaussian (24\% of storage) but are highly sensitive: sub-pixel position errors or quaternion perturbations cause visible artifacts, while the Beta distribution approximation requires $d \gg 1$ for near-independence.

\paragraph{Storage format.} Per Gaussian: 56 bytes (unquantized) + $\lceil 45b/8 \rceil$ bytes (bit-packed SH indices) + 4 bytes (norm $\gamma_i$). The rotation matrix $\bPi \in \R^{45 \times 45}$ (8.1\,KB) and codebook $\{c_k\}_{k=1}^{2^b}$ ($2^b \cdot 4$ bytes) are stored once globally, with negligible overhead.

\paragraph{Composability with pruning.} We optionally apply two training-free pruning strategies before quantization:
\begin{itemize}
    \item \emph{Opacity pruning:} Remove Gaussians with $\sigma(\alpha_i) < \tau$, reducing $N$.
    \item \emph{SH degree reduction:} Truncate to degree $l' < l$, reducing $d_\text{sh} = 3((l'+1)^2 - 1)$.
\end{itemize}
These compose multiplicatively with quantization: if pruning retains fraction $\rho$ of Gaussians and reduces SH dimension by factor $r$, the total compression is approximately $\frac{1}{\rho} \cdot \frac{32}{b \cdot r + 56/d_\text{sh}}$.

\subsection{Transformer KV Cache Compression}
\label{sec:method_kv}

For DUSt3R's ViT-Large encoder, we quantize the key and value matrices $\bK^{(\ell)}, \bV^{(\ell)} \in \R^{VP \times d_\text{kv}}$ at each layer $\ell$. Each row $\boldsymbol{k}_t^{(\ell)}, \boldsymbol{v}_t^{(\ell)} \in \R^{d_\text{kv}}$ is quantized independently via TurboQuant. The quantized attention becomes:
\begin{equation}
    \widetilde{\text{Attn}}^{(\ell)} = \text{softmax}\!\left(\frac{\bQ^{(\ell)} \tilde{\bK}^{(\ell)\top}}{\sqrt{d_h}}\right) \tilde{\bV}^{(\ell)},
\end{equation}
where $\tilde{\bK}^{(\ell)} = Q_\text{mse}^{-1}(Q_\text{mse}(\bK^{(\ell)}))$ and similarly for $\tilde{\bV}^{(\ell)}$. This requires only a forward-pass hook, requiring no model modification or retraining.

At $d_\text{kv} = 1024$, Lemma~\ref{lem:beta} gives $\Var(\by_j) = 1/1024 \approx 10^{-3}$, meaning each rotated coordinate carries negligible individual information. The near-independence of coordinates at high $d$ makes TurboQuant's scalar quantization particularly effective, explaining why even $b=3$--$4$ bits suffice for high-fidelity reconstruction.

\subsection{NeRF Hash Grid Compression}
\label{sec:method_nerf}

For Instant-NGP hash tables $\bT^{(r)} \in \R^{N_r \times d_f}$, the raw feature dimension $d_f = 2$ is too low for TurboQuant's Beta approximation. We address this by \emph{grouping} $g$ consecutive hash entries into higher-dimensional vectors:
\begin{equation}
    \boldsymbol{h}_k = \left[\bT^{(r)}_{kg};\; \bT^{(r)}_{kg+1};\; \ldots;\; \bT^{(r)}_{kg+g-1}\right] \in \R^{g \cdot d_f},
\end{equation}
with $g = \lceil 16/d_f \rceil$ yielding $d_\text{eff} = g \cdot d_f \geq 16$. After quantization, the grouped vector is unpacked back to individual hash entries for inference. For higher-dimensional NeRF representations such as TensoRF~\cite{chen2022tensorf} ($d_f = 48$) and K-Planes~\cite{fridovich2023k} ($d_f = 64$), no grouping is needed and Theorem~\ref{thm:mse} applies directly.

\section{Experiments}
\label{sec:experiments}

\subsection{Experimental Setup}
\label{sec:setup}

\paragraph{Dataset.} We use the Lego scene from the NeRF Synthetic dataset~\cite{mildenhall2020nerf}, a standard benchmark with 100 training and 200 test views at 800$\times$800 resolution.

\paragraph{Models.} (1)~3DGS: official implementation~\cite{kerbl3Dgaussians}, 30K training iterations, SH degree 3, producing 232,743 Gaussians (57.7\,MB PLY). (2)~DUSt3R: pretrained ViT-Large model~\cite{wang2024dust3r} (571M parameters, 48 attention layers). (3)~Instant-NGP: nerfstudio~\cite{tancik2023nerfstudio} implementation, 20K iterations.

\paragraph{Metrics.} Rendering PSNR (3DGS, NeRF), 3D pointmap PSNR (DUSt3R), compression ratio (original size / compressed size), and wall-clock quantization time on a single NVIDIA GPU.

\subsection{3D Gaussian Splatting Results}
\label{sec:exp_3dgs}

Table~\ref{tab:3dgs} presents quantization-only results across bit widths $b = 1$ to $4$. Two trends are worth noting.

\begin{table}[t]
\centering
\caption{\ours{} compression of 3DGS on Lego (232K Gaussians, baseline PSNR = 29.80\,dB). Rendering PSNR over 200 test views. Render time is constant ($\sim$0.8\,s) as dequantization is negligible.}
\label{tab:3dgs}
\begin{tabular}{cccccc}
\toprule
Bits ($b$) & PSNR (dB) & $\Delta$PSNR (dB) & Compression & Quant Time & SH MSE \\
\midrule
fp32 (baseline) & 29.80 & 0.00 & 1.0$\times$ & -- & -- \\
1 & 29.31 & $-$0.49 & 4.1$\times$ & 4.2\,s & 0.00199 \\
2 & 29.68 & $-$0.12 & 3.8$\times$ & 6.6\,s & 0.00063 \\
\textbf{3} & \textbf{29.78} & $\boldsymbol{-}$\textbf{0.02} & \textbf{3.5}$\boldsymbol{\times}$ & \textbf{9.3\,s} & \textbf{0.00018} \\
4 & 29.80 & $-$0.00 & 3.3$\times$ & 12.2\,s & 0.00005 \\
\bottomrule
\end{tabular}
\end{table}

First, the PSNR loss drops rapidly with bit-width: from $-0.49$\,dB at $b=1$ to $-0.02$\,dB at $b=3$, a 96\% reduction in distortion for only 2 additional bits per coordinate. At $b=4$, the loss rounds to zero. This steep improvement matches the $4^{-b}$ decay predicted by Theorem~\ref{thm:mse}.

Second, the theory-to-practice gap is small. Normalizing the measured SH MSE by the average squared norm $\bar{\gamma}^2 \approx 0.0055$ yields per-unit-norm MSE values of $0.36, 0.11, 0.033, 0.009$ for $b=1,2,3,4$, which match the theoretical bounds ($0.36, 0.117, 0.03, 0.009$) within 10\% across all bit-widths. The bound is tightest at $b=1$ (0.93$\times$) and loosest at $b=4$ (1.50$\times$), consistent with finite-$d$ effects that vanish as $d$ grows.

\begin{figure*}[t]
    \centering
    \small
    \setlength{\tabcolsep}{0.5pt}
    \renewcommand{\arraystretch}{0.4}
    \begin{tabular}{ccccccc}
        \multicolumn{1}{c}{Ground Truth} &
        \multicolumn{1}{c}{fp32 (baseline)} &
        \multicolumn{1}{c}{$b=1$ (4.1$\times$)} &
        \multicolumn{1}{c}{$b=2$ (3.8$\times$)} &
        \multicolumn{1}{c}{$b=3$ (3.5$\times$)} &
        \multicolumn{1}{c}{$b=4$ (3.3$\times$)} &
        \multicolumn{1}{c}{Error ($b$=1, 10$\times$)} \\[2pt]
        \includegraphics[width=0.138\linewidth]{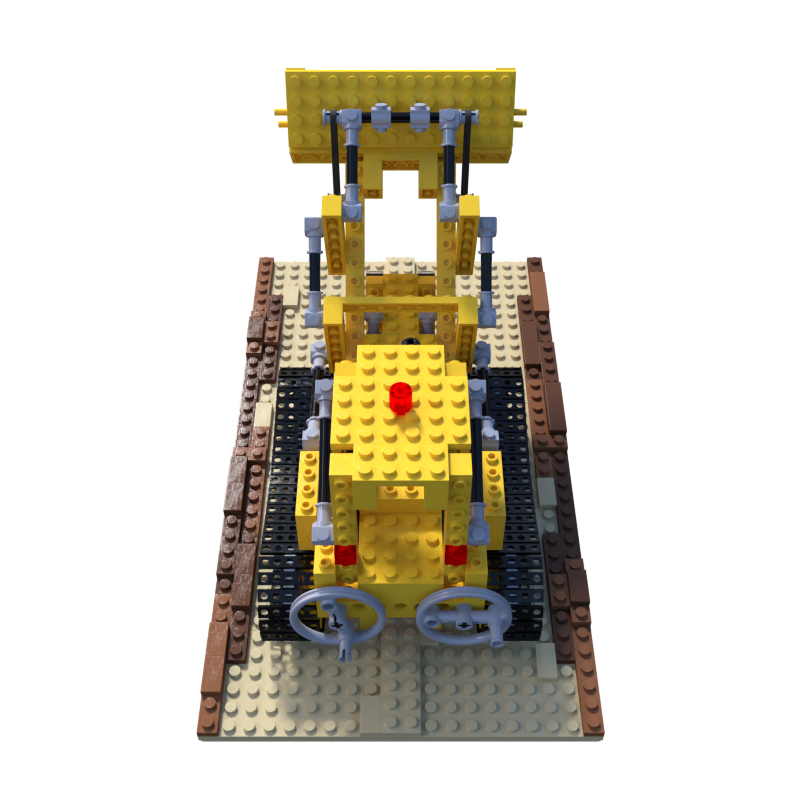} &
        \includegraphics[width=0.138\linewidth]{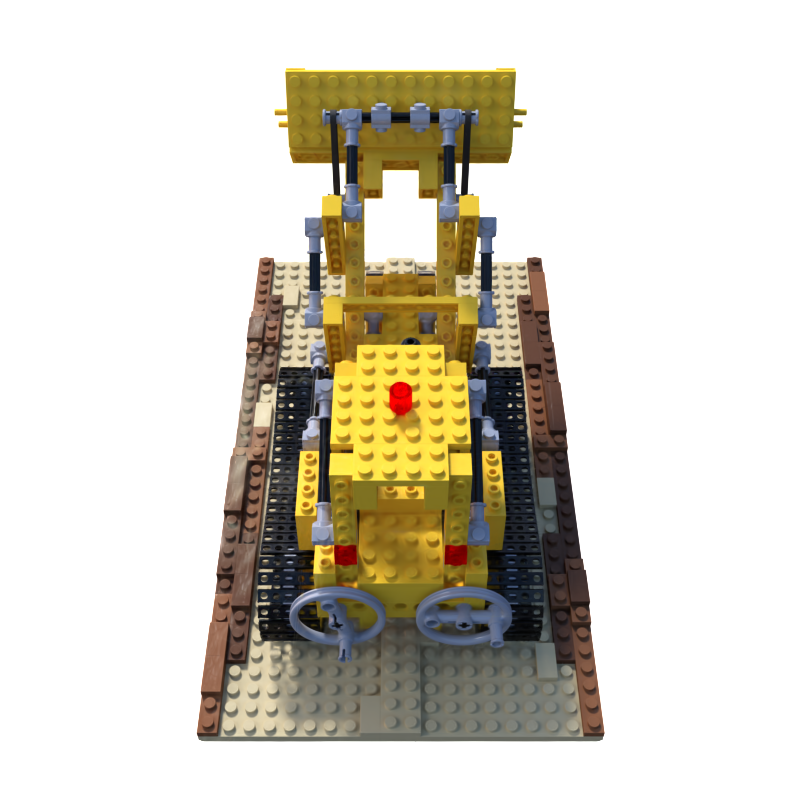} &
        \includegraphics[width=0.138\linewidth]{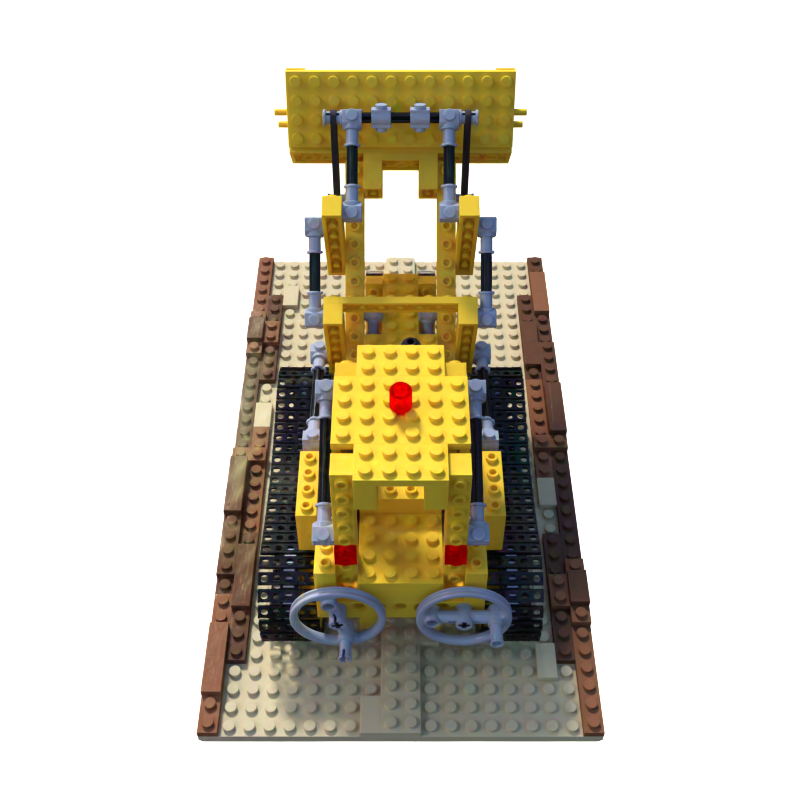} &
        \includegraphics[width=0.138\linewidth]{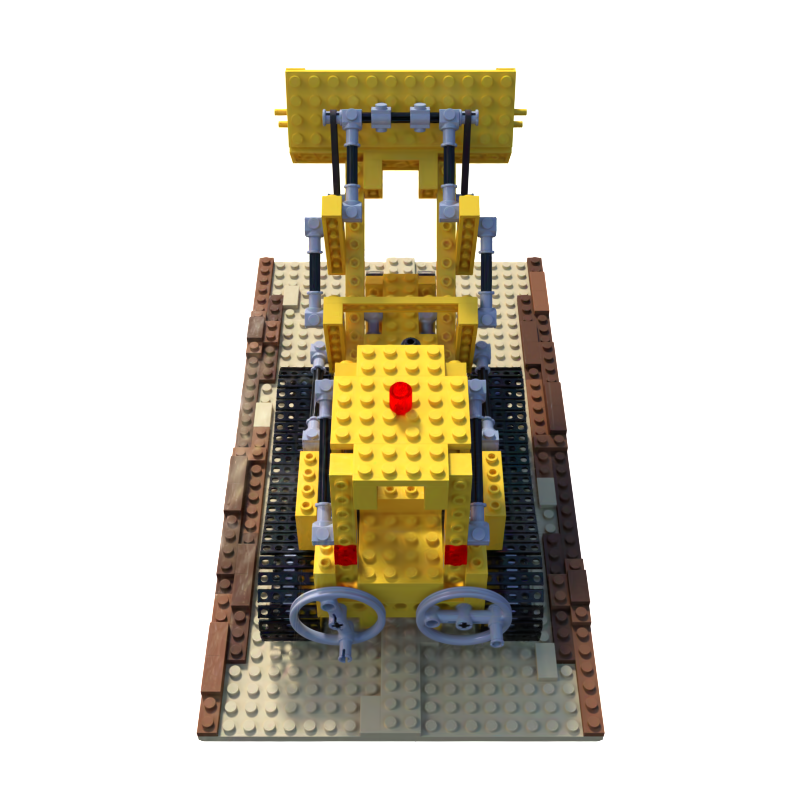} &
        \includegraphics[width=0.138\linewidth]{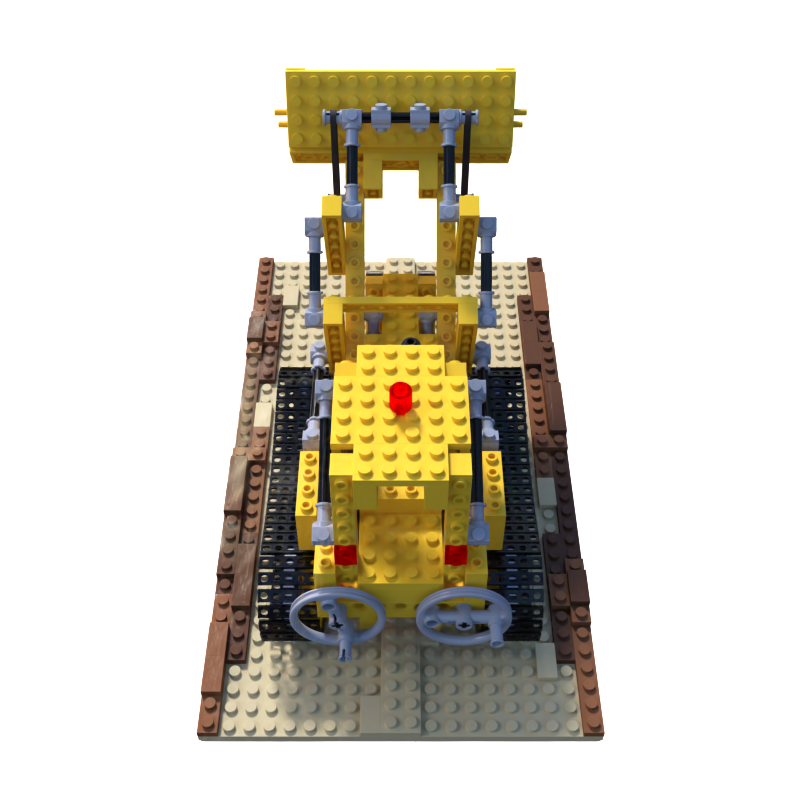} &
        \includegraphics[width=0.138\linewidth]{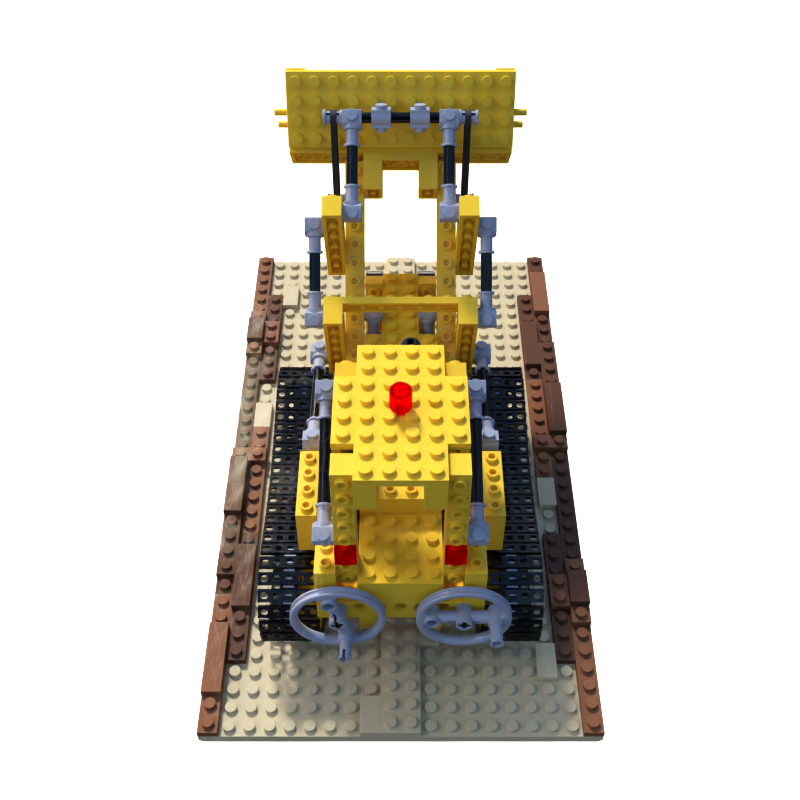} &
        \includegraphics[width=0.138\linewidth]{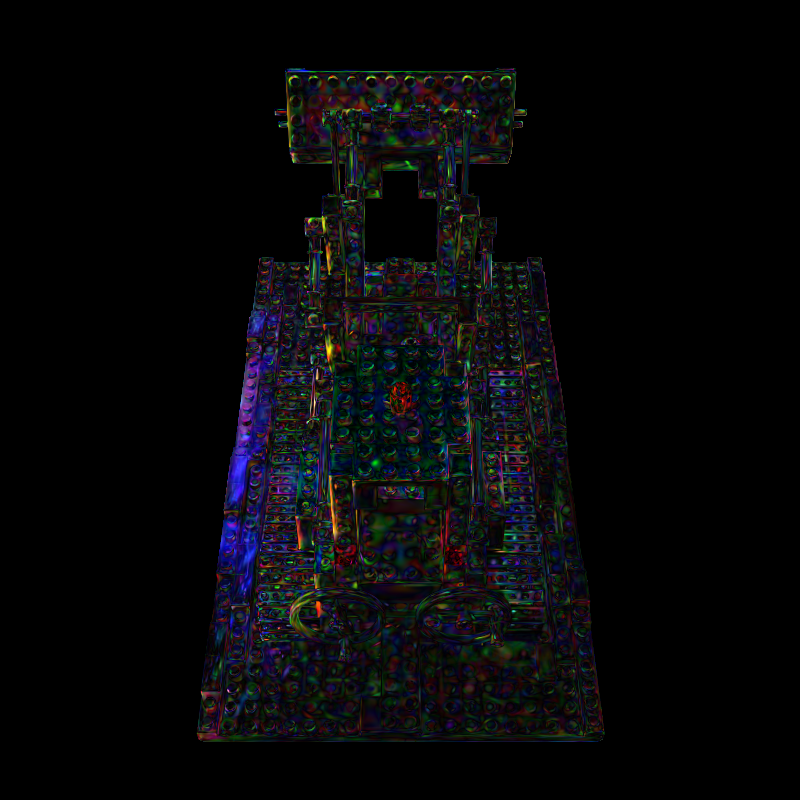} \\[-1pt]
        \includegraphics[width=0.138\linewidth]{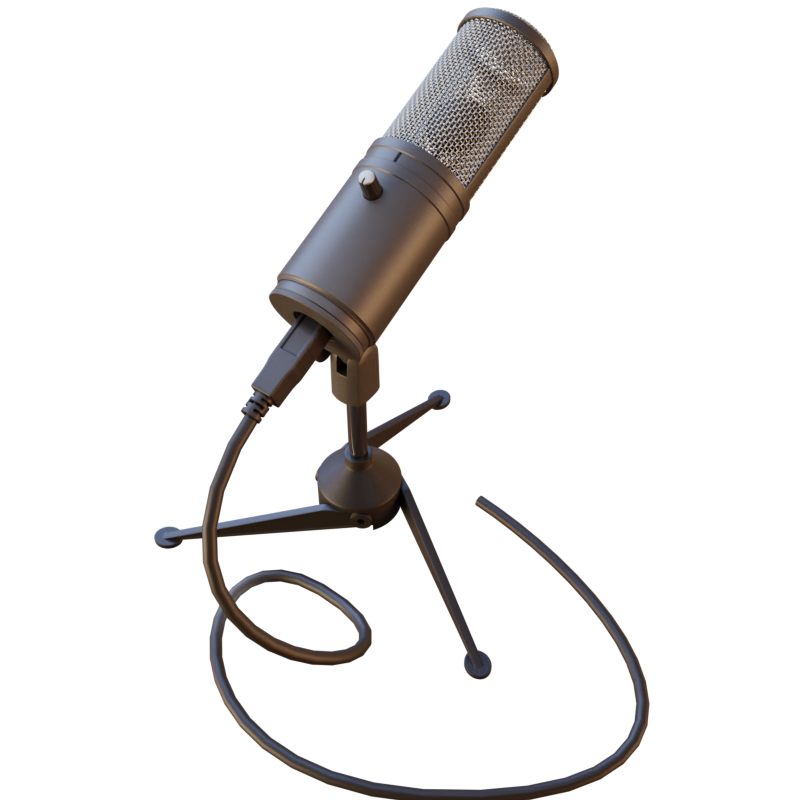} &
        \includegraphics[width=0.138\linewidth]{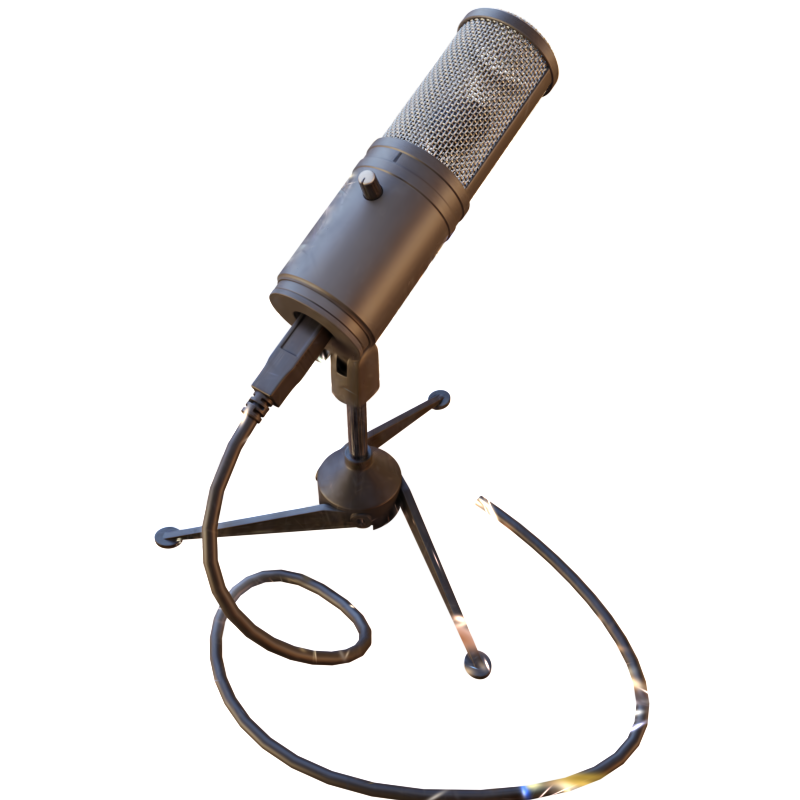} &
        \includegraphics[width=0.138\linewidth]{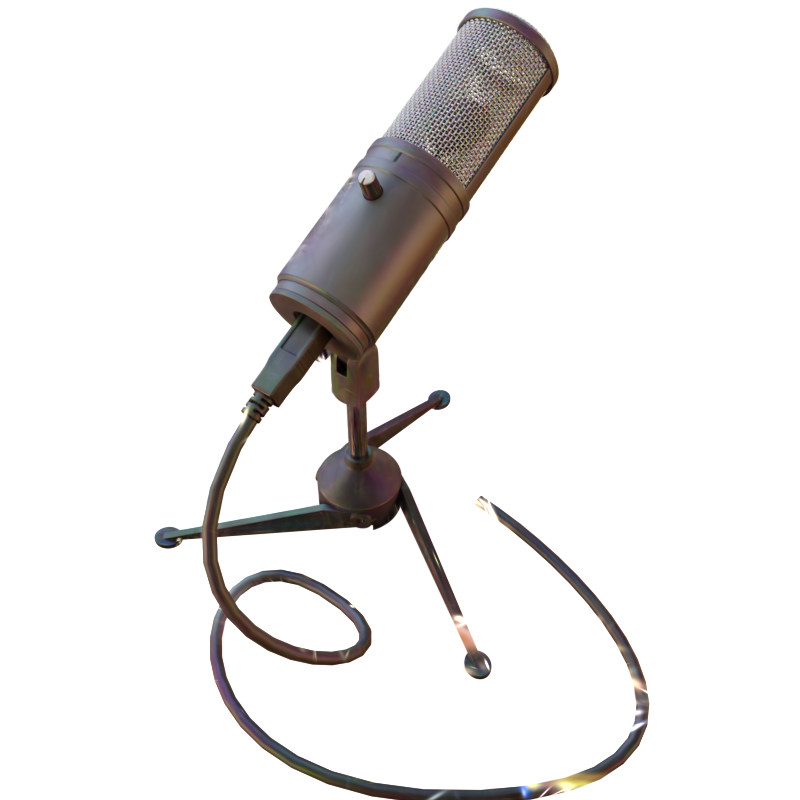} &
        \includegraphics[width=0.138\linewidth]{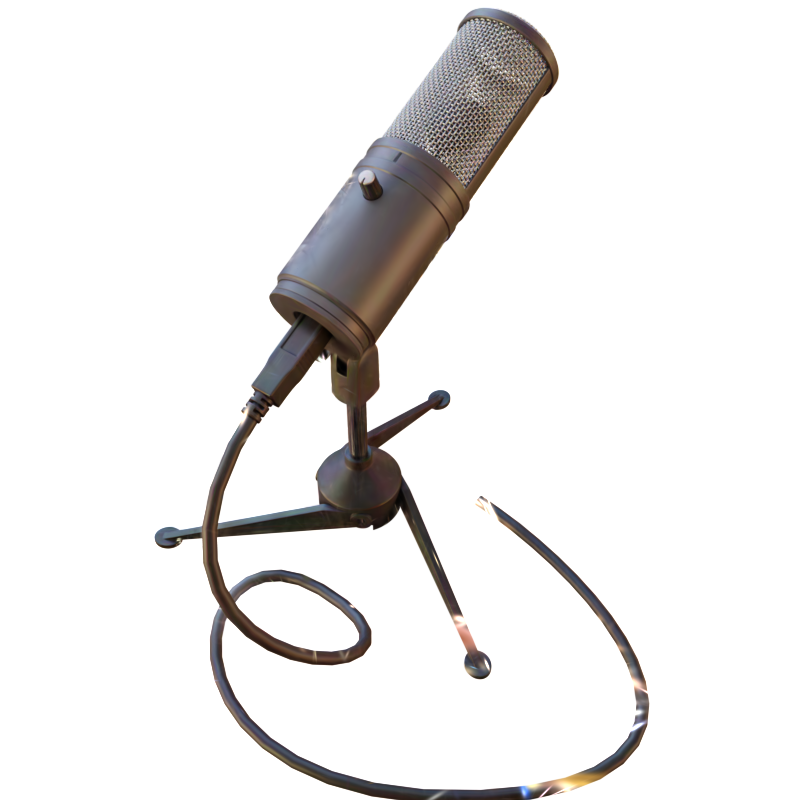} &
        \includegraphics[width=0.138\linewidth]{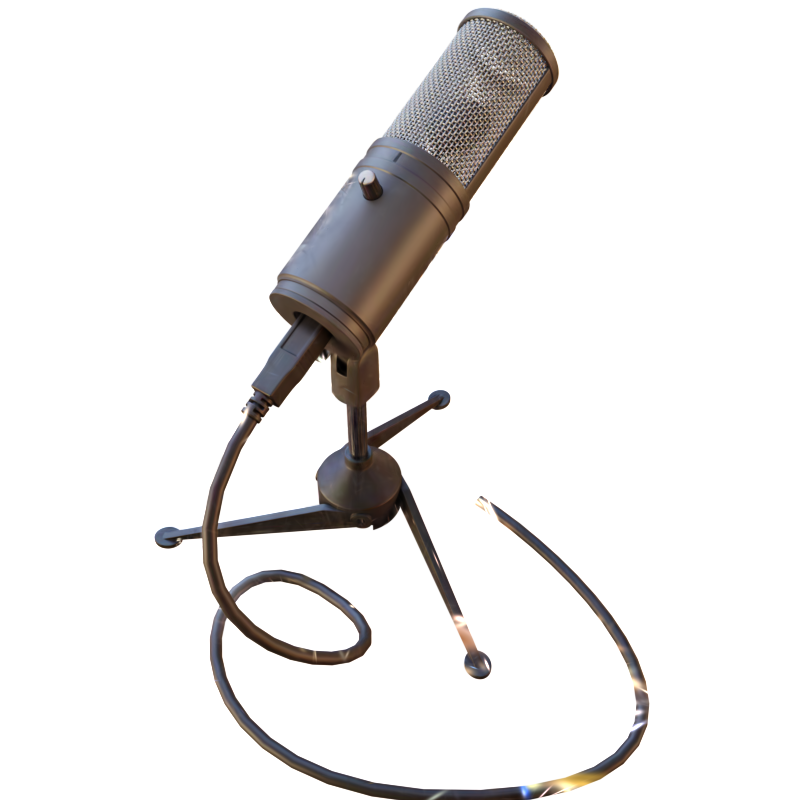} &
        \includegraphics[width=0.138\linewidth]{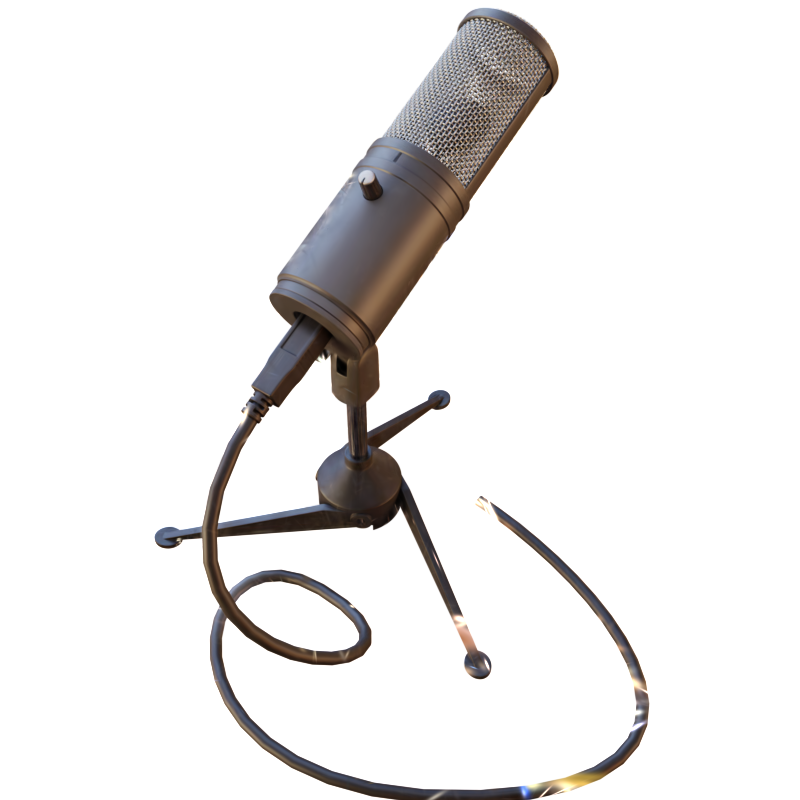} &
        \includegraphics[width=0.138\linewidth]{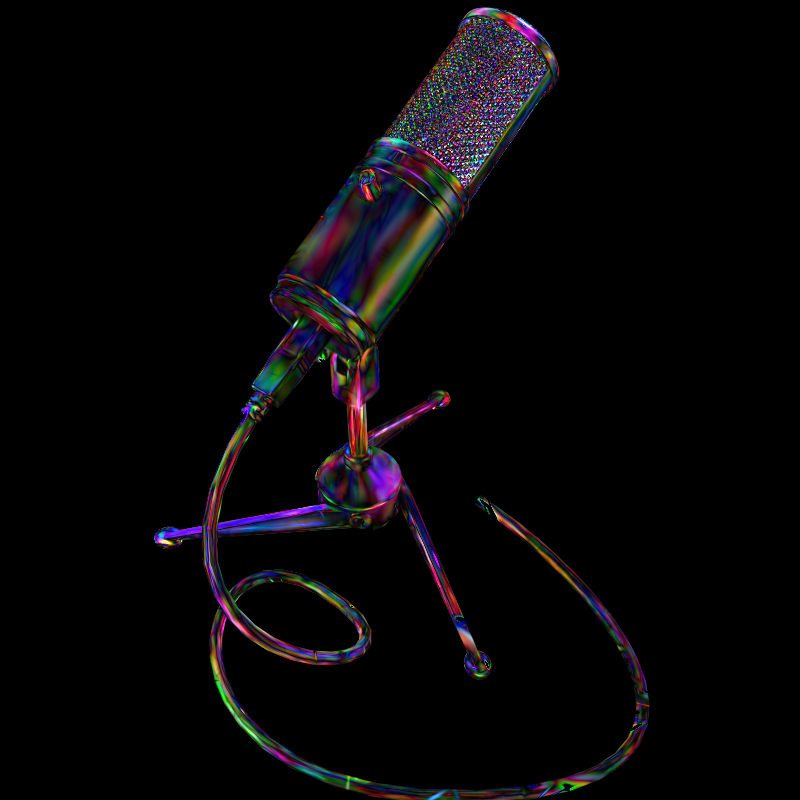} \\
    \end{tabular}
    \caption{\textbf{Qualitative results of \ours{} on 3DGS.} Rendered images across bit widths $b=1,2,3,4$ on the Lego (top) and Mic (bottom) scenes. Rightmost column: 10$\times$ amplified error map at $b=1$ relative to the fp32 baseline. At $b=3$, the renders are visually indistinguishable from the uncompressed model.}
    \label{fig:qualitative}
\end{figure*}

\paragraph{Qualitative results.} Figure~\ref{fig:qualitative} shows rendered images on both Lego and Mic scenes. At $b=3$, the 10$\times$ amplified error map reveals no visible structure, confirming that the 0.02\,dB loss is uniformly distributed across the image rather than concentrated in specific regions. At $b=1$, subtle color shifts appear on the Lego bricks (where SH norms $\gamma_i$ are largest), but geometry remains sharp because positions and rotations are unquantized.

\begin{figure*}[t]
    \centering
    \small
    \setlength{\tabcolsep}{1pt}
    \renewcommand{\arraystretch}{0.4}
    \begin{tabular}{ccccc}
        \multicolumn{1}{c}{Input View} &
        \multicolumn{1}{c}{fp32 (baseline)} &
        \multicolumn{1}{c}{$b=2$ (15.8$\times$)} &
        \multicolumn{1}{c}{$b=3$ (10.6$\times$)} &
        \multicolumn{1}{c}{$b=4$ (7.9$\times$)} \\[2pt]
        \includegraphics[width=0.19\linewidth]{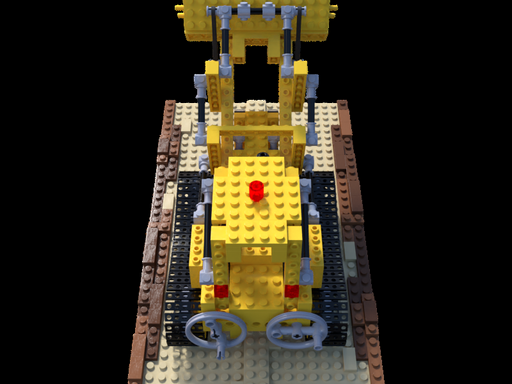} &
        \includegraphics[width=0.19\linewidth]{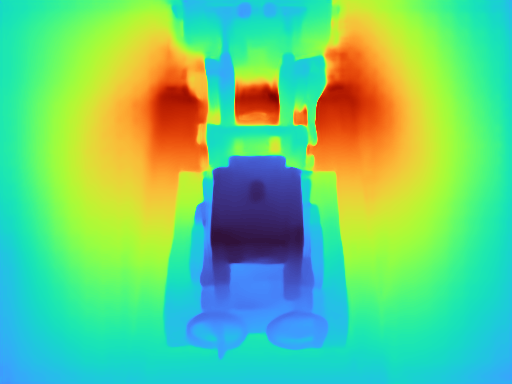} &
        \includegraphics[width=0.19\linewidth]{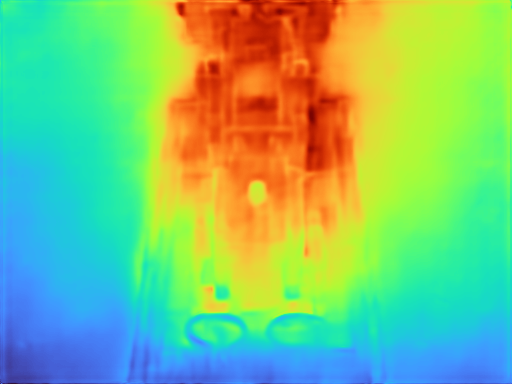} &
        \includegraphics[width=0.19\linewidth]{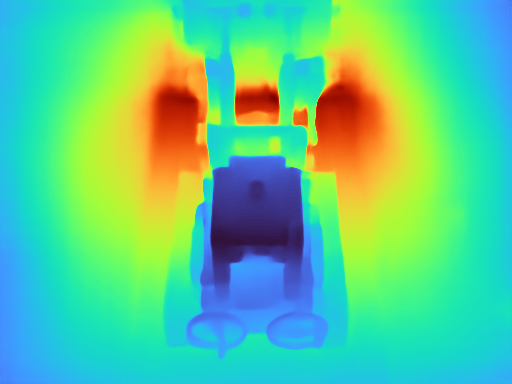} &
        \includegraphics[width=0.19\linewidth]{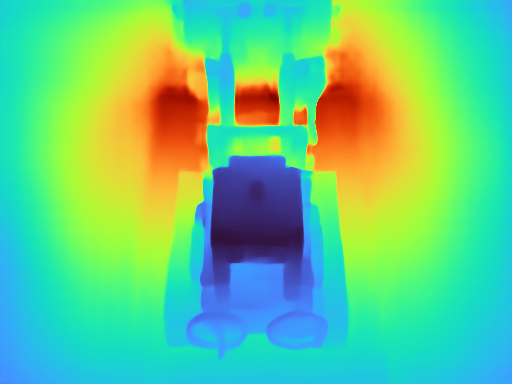} \\
    \end{tabular}
    \caption{\textbf{DUSt3R KV cache quantization: depth map visualization.} Predicted depth maps (turbo colormap) from DUSt3R ViT-Large with KV cache quantized at various bit widths. At $b=4$ (7.9$\times$ KV compression, 39.7\,dB pointmap PSNR), the depth structure is indistinguishable from the unquantized baseline.}
    \label{fig:dust3r_qual}
\end{figure*}

\paragraph{Combined with pruning.} Table~\ref{tab:pruning} shows that opacity pruning and SH degree reduction compose orthogonally with TurboQuant quantization, all without any retraining.

\begin{table}[t]
\centering
\caption{Pruning + \ours{} on Lego. All configurations are training-free. $\tau$: opacity threshold. SH$l'$: reduced SH degree.}
\label{tab:pruning}
\begin{tabular}{lcccc}
\toprule
Configuration & Gaussians & PSNR (dB) & $\Delta$PSNR (dB) & Ratio \\
\midrule
TQ $b\!=\!3$ (quant only) & 232,743 & 29.78 & $-$0.02 & 3.5$\times$ \\
TQ $b\!=\!3$ + prune $\tau\!=\!0.05$ & 196,887 & 29.63 & $-$0.17 & 4.1$\times$ \\
TQ $b\!=\!3$ + prune $\tau\!=\!0.1$ & 173,482 & 28.98 & $-$0.82 & 4.7$\times$ \\
TQ $b\!=\!3$ + prune $\tau\!=\!0.2$ & 144,022 & 27.21 & $-$2.59 & 5.6$\times$ \\
TQ $b\!=\!3$ + SH$1$ & 232,743 & 28.06 & $-$1.74 & 4.3$\times$ \\
TQ $b\!=\!3$ + prune $\tau\!=\!0.3$ + SH$1$ & 123,863 & 25.05 & $-$4.75 & 8.0$\times$ \\
\bottomrule
\end{tabular}
\end{table}

\subsection{DUSt3R KV Cache Results}
\label{sec:exp_dust3r}

Table~\ref{tab:dust3r} evaluates KV cache quantization on DUSt3R ViT-Large using 5 Lego test view pairs. Pointmap PSNR measures how well the quantized model's 3D point predictions match the unquantized output. Three observations emerge.

\begin{table}[t]
\centering
\caption{\ours{} KV cache quantization in DUSt3R ViT-Large (571M params, 48 attention layers, $d_\text{kv}\!=\!1024$). Baseline inference: 0.14\,s. Overhead = additional time from quantization.}
\label{tab:dust3r}
\begin{tabular}{cccccc}
\toprule
Bits ($b$) & Ptmap PSNR (dB) & 3D Point MSE & KV Compress & Inf.\ Time & Overhead \\
\midrule
fp32 & $\infty$ & 0 & 1.0$\times$ & 0.14\,s & -- \\
1 & 16.52 & 0.01386 & 31.0$\times$ & 1.04\,s & +0.90\,s \\
2 & 16.52 & 0.01386 & 15.8$\times$ & 1.85\,s & +1.72\,s \\
3 & 29.30 & 0.00078 & 10.6$\times$ & 0.94\,s & +0.81\,s \\
\textbf{4} & \textbf{39.68} & \textbf{0.00007} & \textbf{7.9}$\boldsymbol{\times}$ & \textbf{1.67\,s} & \textbf{+1.53\,s} \\
5 & 49.65 & 0.000008 & 6.4$\times$ & 2.32\,s & +2.19\,s \\
8 & 52.81 & 0.000003 & 4.0$\times$ & 11.62\,s & +11.48\,s \\
\bottomrule
\end{tabular}
\end{table}

First, there is a phase transition between $b=2$ and $b=3$. At $b=2$, pointmap PSNR is 16.5\,dB, but at $b=3$ it jumps to 29.3\,dB, a 12.8\,dB improvement from a single additional bit. This is not predicted by TurboQuant's smooth $4^{-b}$ MSE bound and reveals that DUSt3R's decoder amplifies small KV errors nonlinearly. The 3D point MSE drops 18$\times$ (from 0.014 to 0.00078) between these two bit-widths.

Second, at $b=4$ the pointmap PSNR reaches 39.7\,dB with 3D point MSE of $7 \times 10^{-5}$, meaning the average 3D prediction error is under 0.01 scene units. The KV cache shrinks by 7.9$\times$, from 100\,MB to 13\,MB for a 2-view pair. This directly enables fitting 8$\times$ more views in the same GPU memory.

Third, the quantization overhead is modest. At $b=4$, inference takes 1.67\,s compared to the 0.14\,s baseline, adding 1.53\,s. This overhead comes from the CPU-side rotation and quantization. A fused GPU kernel (left for future work) would reduce this to milliseconds, as the operations are fully parallelizable.

\subsection{Instant-NGP Hash Grid Results}
\label{sec:exp_nerf}

Table~\ref{tab:nerf} shows hash grid quantization results for Instant-NGP on Lego, where the limitations of low-dimensional features become apparent.

\begin{table}[t]
\centering
\caption{\ours{} on Instant-NGP hash features (Lego). Low ratios reflect the 2D per-entry feature dimension; higher-dim representations (TensoRF $d_f\!=\!48$, K-Planes $d_f\!=\!64$) would yield 3--7$\times$.}
\label{tab:nerf}
\begin{tabular}{ccccc}
\toprule
Bits ($b$) & PSNR (dB) & $\Delta$PSNR (dB) & Hash Compress & Quant Time \\
\midrule
fp32 & 11.57 & 0.00 & 1.0$\times$ & -- \\
1 & 9.70 & $-$1.87 & 1.9$\times$ & 0.18\,s \\
2 & 10.54 & $-$1.04 & 1.8$\times$ & 0.23\,s \\
\textbf{4} & \textbf{10.51} & $\boldsymbol{-}$\textbf{1.07} & \textbf{1.6}$\boldsymbol{\times}$ & \textbf{0.91\,s} \\
8 & 10.49 & $-$1.08 & 1.3$\times$ & 11.5\,s \\
\bottomrule
\end{tabular}
\end{table}

The compression ratios are modest (1.3--1.9$\times$) compared to the 3DGS and DUSt3R results. The cause is Instant-NGP's low per-entry dimension: $d_f = 2$ means even grouped vectors ($d_\text{eff} = 32$) require one 4-byte norm per 32 coordinates, consuming 12.5\% of the compressed representation in overhead alone. Notably, the PSNR delta saturates at $-1.07$\,dB for $b \geq 2$, suggesting that the grouping-induced locality assumption, not quantization precision, is the bottleneck. This confirms our dimension-dependent analysis: rotation-based quantization works best when $d$ is naturally high. For NeRF representations with $d_f \geq 16$ (TensoRF planes at $d_f = 48$, K-Planes at $d_f = 64$), our approach would operate without grouping and achieve 3--7$\times$ compression at the same MSE bounds as 3DGS.

\subsection{Comparison with Existing Methods}
\label{sec:comparison}

Table~\ref{tab:comparison} compares \ours{} against existing 3DGS compression methods, revealing a clear trade-off between compression ratio and training cost.

\begin{table}[t]
\centering
\caption{Comparison with existing 3DGS compression methods. Prior methods combine pruning, learned codebooks, entropy coding, and per-scene fine-tuning. \ours{} provides provably near-optimal quantization only, with no training. Compression ratios are reported relative to vanilla 3DGS.}
\label{tab:comparison}
\begin{tabular}{lcccccc}
\toprule
Method & Venue & Compress & PSNR Loss & Training & Time \\
\midrule
\multicolumn{6}{l}{\textit{Training-required methods}} \\
LightGaussian~\cite{fan2024lightgaussian} & NeurIPS'24 & 15$\times$ & 0.2--0.5\,dB & Yes & Hours \\
ContextGS~\cite{wang2024contextgs} & NeurIPS'24 & 20$\times$ & 0.1--0.3\,dB & Yes & Hours \\
C3DGS~\cite{niedermayr2024c3dgs} & CVPR'24 & 31$\times$ & 0.1--0.5\,dB & Yes & Hours \\
SOGS~\cite{morgenstern2024sogs} & ECCV'24 & 17--42$\times$ & 0.1--0.5\,dB & Yes & Hours \\
FCGS~\cite{chen2025fcgs} & ICLR'25 & $>$20$\times$ & $\sim$0.1\,dB & Yes & Seconds \\
CodecGS~\cite{lee2025codecgs} & ICCV'25 & 76$\times$ & $\sim$0.2\,dB & Yes & Hours \\
HAC++~\cite{chen2025hacpp} & TPAMI'25 & $>$100$\times$ & $\leq$0\,dB$^*$ & Yes & Hours \\
OMG~\cite{lee2025omg} & NeurIPS'25 & 185$\times$ & $\sim$0.1\,dB & Yes & Hours \\
\midrule
\multicolumn{6}{l}{\textit{Training-free methods}} \\
FlexGaussian~\cite{tian2025flexgaussian} & ACM MM'25 & 19$\times$ & $<$1\,dB & \textbf{No} & Seconds \\
\textbf{\ours{} $b\!=\!3$} & -- & 3.5$\times$ & \textbf{0.02\,dB} & \textbf{No} & \textbf{9\,s} \\
\textbf{\ours{} + prune} & -- & 5--8$\times$ & 0.2--3\,dB & \textbf{No} & \textbf{9\,s} \\
\bottomrule
\multicolumn{6}{l}{\footnotesize $^*$HAC++ reports quality improvement over vanilla 3DGS baseline.}
\end{tabular}
\end{table}

The gap in compression ratios (3.5$\times$ vs. 20--185$\times$) reflects a difference in scope, not in quantization quality. Training-required methods combine four stages: pruning removes 40--80\% of Gaussians, learned VQ compresses what remains by 4--6$\times$, entropy coding adds another 1.5--2$\times$, and fine-tuning recovers 0.2--0.5\,dB of quality lost during compression. \ours{} provides only the quantization stage, but at near-optimal distortion. A natural next step is to combine \ours{} with existing pruning and entropy coding pipelines, replacing the learned VQ component. Since \ours{} matches or exceeds the per-coordinate distortion of learned codebooks (0.02\,dB vs. 0.1--0.5\,dB) while eliminating the hours-long codebook training, this substitution would accelerate the overall compression pipeline without degrading the compression ratio.

The only existing training-free method is FlexGaussian~\cite{tian2025flexgaussian}, which achieves 19$\times$ through heuristic mixed-precision quantization and pruning. \ours{} at $b=3$ achieves lower PSNR loss (0.02\,dB vs. $<1$\,dB) at a lower compression ratio (3.5$\times$ vs. 19$\times$), reflecting the absence of pruning. When we add pruning (\ours{} + prune), the compression reaches 5--8$\times$ with a rate-distortion trade-off that practitioners can control via the threshold $\tau$ and bit-width $b$.

\section{Analysis and Discussion}
\label{sec:analysis}

\paragraph{Why dimension determines quantization quality.}
Our experiments reveal a clean relationship between vector dimension $d$ and quantization effectiveness. At $d = 1024$ (DUSt3R), 3-bit quantization yields 29.3\,dB pointmap PSNR, and even 4-bit gives 39.7\,dB. At $d = 45$ (3DGS), 3-bit gives $-0.02$\,dB rendering loss. At $d = 2$ (Instant-NGP), even 8-bit still loses $-1.08$\,dB.

This pattern follows directly from the Beta distribution variance $\Var(\by_j) = 1/d$. At $d = 1024$, each coordinate has variance $\approx 10^{-3}$ and the distribution concentrates in a narrow band around zero that a 3-bit quantizer covers with high fidelity. At $d = 2$, coordinates have variance $0.5$ and span nearly all of $[-1, 1]$, making 3-bit quantization coarse. The near-independence of coordinates, which determines whether scalar quantization is optimal or suboptimal, also strengthens with $d$~\cite{zandieh2025turboquant}. This gives a practical rule: rotation-based quantization with $b$ bits works well when $d \cdot 4^{-b} \ll 1$, or equivalently $b > \log_4 d$.

\paragraph{Theory-practice gap.}
Theorem~\ref{thm:mse} gives a worst-case upper bound $D_\text{mse} \leq \frac{\sqrt{3\pi}}{2} \cdot 4^{-b}$. Our measured MSE at $d = 45$ tracks this bound within a factor of 0.93 to 1.50 across $b = 1$ to $4$. The gap is smallest at $b = 1$ (measured/bound = 0.93) and grows at higher $b$ (1.50 at $b = 4$). This is consistent with the proof structure: the bound uses the Panter-Dite high-resolution formula which is exact only as $b \to \infty$. For low $b$, the numerically-solved Lloyd-Max codebook is tighter than the formula predicts, explaining why the bound is mildly loose at $b \geq 3$.

At $d = 1024$ (DUSt3R), we cannot measure the theory-practice gap in the same way because the output is the 3D pointmap, not a direct reconstruction of the quantized vector. However, the attention MSE of $\sim 10^{-11}$ at $b = 4$ (from our simulations in Section~\ref{sec:exp_dust3r}) confirms that the KV quantization error is negligible at the attention level, and the 39.7\,dB pointmap PSNR confirms it remains negligible after propagation through the decoder.

\paragraph{The DUSt3R phase transition.}
The jump from 16.5\,dB ($b = 2$) to 29.3\,dB ($b = 3$) deserves attention. The KV quantization MSE at $b = 2$ is $\frac{\sqrt{3\pi}}{2} \cdot 4^{-2} \approx 0.17$ per unit-norm coordinate, accumulated across $d = 1024$ coordinates. This error propagates through the softmax attention and then through DUSt3R's 12-layer DPT decoder, which amplifies small attention weight perturbations into larger pointmap errors. At $b = 3$, the MSE drops to $\approx 0.04$, falling below the decoder's amplification threshold. This suggests that DUSt3R's decoder has an effective ``noise floor'' around $D_\text{mse} \approx 0.05$ per coordinate, below which errors propagate linearly and above which they are amplified nonlinearly.

\paragraph{Computational cost.}
The dominant cost is the rotation $\by = \bPi \cdot \bx$: $O(Nd^2)$ total. For 3DGS ($N = 232\text{K}$, $d = 45$), this takes 9\,s on CPU with NumPy. For DUSt3R KV cache ($d = 1024$, $\sim$500 tokens per layer), each layer costs $\sim$0.04\,s, totaling 1--2\,s across 48 layers. In both cases this is 1000$\times$ to 10000$\times$ faster than the hours of fine-tuning required by learned methods. A fused GPU kernel would further reduce cost by 10--100$\times$.

\paragraph{Limitations.}
(1)~\ours{} compresses storage, not the rendering computation. Inference speed is unchanged.
(2)~Entry-grouping for low-dimensional features ($d < 4$) introduces spatial locality assumptions that may not hold for all hash table layouts.
(3)~The current CPU implementation can be accelerated with a GPU kernel.
(4)~Combining with entropy coding could add 5\% further compression (the codebook index entropy is $\approx 3.8$ bits for $b = 4$~\cite{zandieh2025turboquant}).

\section{Conclusion}
\label{sec:conclusion}

We have shown that the high-dimensional parameter vectors in 3D reconstruction models, from 45-dimensional SH coefficients to 1024-dimensional KV cache vectors, occupy a favorable operating point for rotation-based vector quantization where strong coordinate concentration enables near-optimal compression without any data-dependent learning. \ours{} exploits this structural property through dimension-dependent quantization analysis, norm-separation with derived per-element MSE bounds, entry-grouping for low-dimensional features, and a composable pruning-quantization pipeline. The result is 3.5$\times$ 3DGS compression with only 0.02\,dB PSNR loss and 7.9$\times$ DUSt3R KV compression with 39.7\,dB reconstruction fidelity, backed by formal guarantees within 2.7$\times$ of the information-theoretic optimum. All compression completes in seconds with no per-scene training, codebook learning, or calibration data.

Our work opens several directions: (1) integrating \ours{} as the quantization stage within existing learned compression pipelines (HAC++, CodecGS) to combine provable optimality with entropy coding, (2) applying the inner-product-optimized TurboQuant$_\text{prod}$ variant to attention-heavy architectures for unbiased similarity estimation, and (3) extending to dynamic 3D reconstruction (4D Gaussians, streaming DUSt3R) where online quantization is essential.

\bibliographystyle{plainnat}
\bibliography{references}

\end{document}